\documentclass[runningheads]{llncs}
\usepackage{graphicx}
\usepackage{comment}
\usepackage{amsmath,amssymb} 
\usepackage{color}
\usepackage{epsfig}

\usepackage[ruled,linesnumbered]{algorithm2e}
\usepackage{subfigure}
\usepackage{mathrsfs}
\usepackage{multirow}
\usepackage{multicol}
\usepackage{xcolor}
\usepackage{times}
\usepackage[american]{babel}

\newcommand{\mca}[1]{\mathcal{#1}}
\newcommand{\mbf}[1]{\mathbf{#1}}
\newcommand{\floor}[1]{\lfloor #1\rfloor}

\newcommand{\smallgap}{\vspace{0.3em}\noindent}
\newcommand{\js}[1]{#1}
\newcommand{\lyx}[1]{#1}
\newcommand{\etal}{\emph{et al.}}

\begin{document}
	\pagestyle{headings}
	\mainmatter
	\def\ECCVSubNumber{2588}  
	
	\title{CFAD: Coarse-to-Fine Action Detector for Spatiotemporal Action Localization} 

	\titlerunning{Coarse-to-Fine Action Detector}
	%
	\author{Yuxi Li\inst{1} \and
	Weiyao Lin\inst{1,2}\thanks{Correspondance author, wylin@sjtu.edu.cn} \and
	John See\inst{3} \and Ning Xu\inst{4} \\ Shugong Xu\inst{2} \and Ke Yan\inst{5} \and Cong Yang\inst{5}}
	\authorrunning{Y. Li et al.}
	%
	\institute{Department of Electronic Engineering, Shanghai Jiao Tong University, China \and
	Institute for Advanced Communication and Data Science, Shanghai University, China \and
	Faculty of Computing and Informatics, Multimedia University, Malaysia \and
	   Adobe Research, USA \and 
	   Clobotics, China
	\\
	}
	\maketitle
	
	\begin{abstract}
		\js{Most current pipelines for spatio-temporal action localization connect frame-wise or clip-wise detection results to generate action proposals,} \lyx{where only local information is exploited and the efficiency is hindered by dense per-frame localization.} In this paper, we propose Coarse-to-Fine Action Detector (CFAD), an original end-to-end trainable framework for \js{efficient} 
		spatio-temporal action localization. 
		The CFAD \js{introduces a new paradigm that} first 
		estimates coarse spatio-temporal action tubes \js{from video streams}, and then refines the tubes' location \js{based} on key timestamps. This \js{concept} is implemented by two key components, the Coarse and Refine Modules in our framework. The \js{parameterized modeling} 
		of long temporal information in the Coarse Module helps obtain accurate \js{initial} tube estimation, while the Refine Module \js{selectively} adjusts the tube location under the guidance of key timestamps. Against other methods, the proposed CFAD achieves competitive results on 
		\js{action}
		detection benchmarks of UCF101-24, UCFSports and JHMDB-21 with \js{inference} speed that is $3.3 \times$ faster than the nearest competitor.
	\end{abstract}

	\section{Introduction}
	
	Spatial-temporal action detection is the task of recognizing actions from input videos and localizing them in space and time. In contrast to action recognition or temporal localization, this task is far more complex, requiring both temporal detection along the time span and spatial detection at each frame when the actions occur.
	
\begin{figure}[tb]
		\centering
		\subfigure[]{\includegraphics[width=0.49\textwidth]{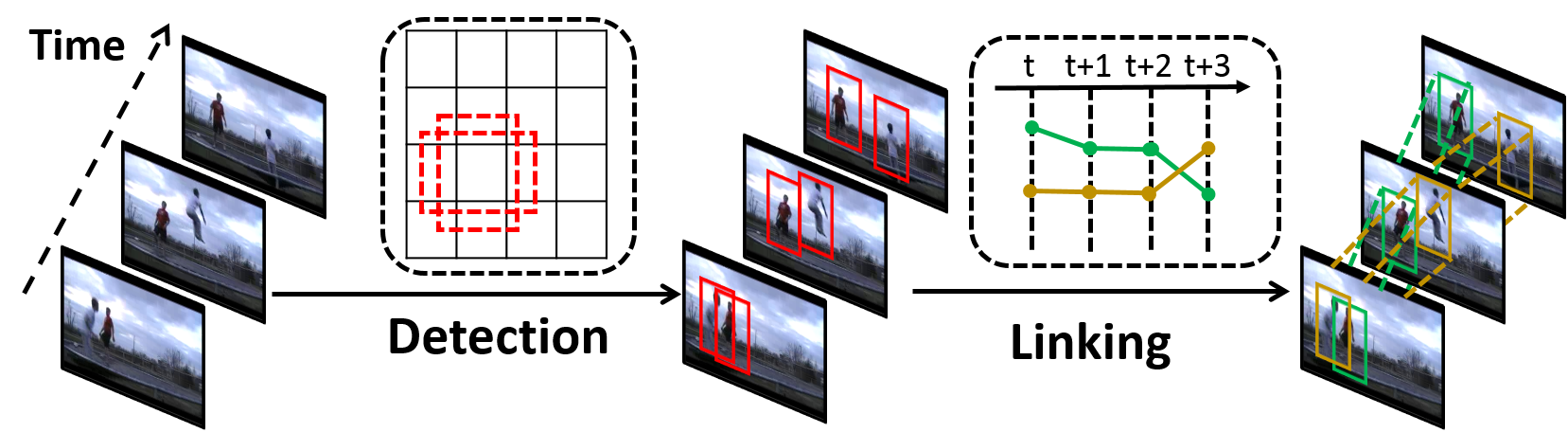}\label{fig:detect-link}}
		\vline{}
		\subfigure[]{\includegraphics[width=0.49\textwidth]{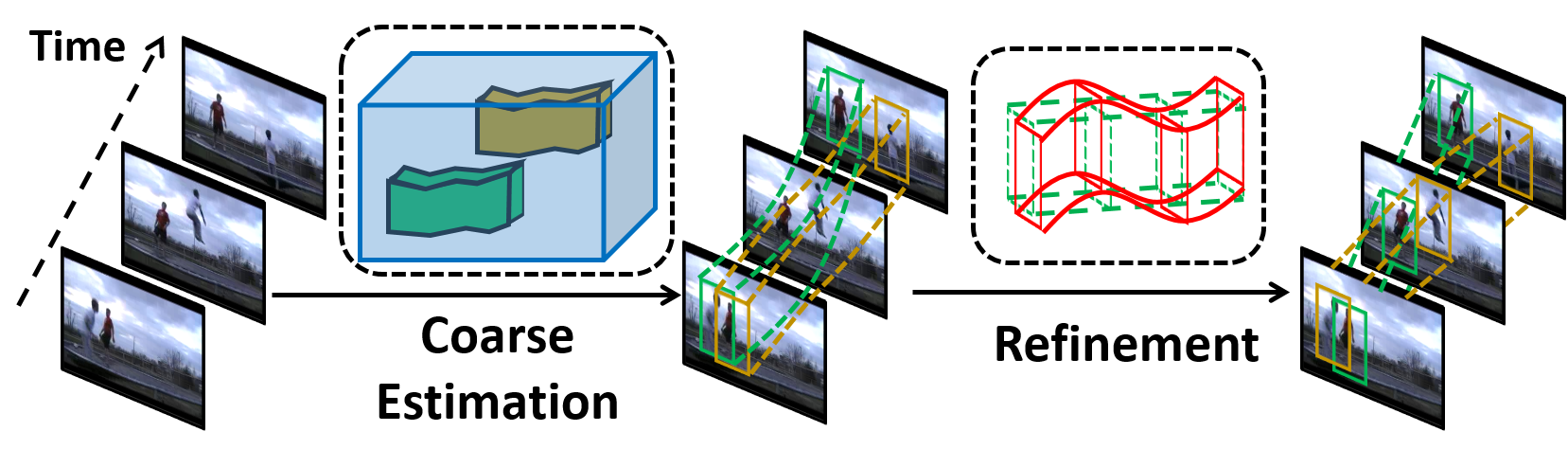}\label{fig:coarse-fine}}
		
		\caption{The comparison between pipelines of detection and linking and our coarse-to-fine framework. (a) workflow of detection and linking method in previous works. (b) Our coarse-to-fine method to detect action tubes. (Best viewed in color.)}
		\label{fig:intro}
	\end{figure}

	Most existing methods for spatiotemporal action detection~\cite{Gkioxari_2015,peng2016multi,saha2016deep,saha2017amtnet,singh2017online,hou2017end,zhao2019dance,yang2019step} are implemented in two stages (illustrated in Fig.~\ref{fig:detect-link}). First, a spatial detector is applied to generate dense action box proposals on each frame. Then, these frame-level detections are linked together by a certain heuristic algorithm to generate final output, which is a series of boxes or an \emph{action tube}. Nevertheless, since these approaches take a single or stack of frames 
	as input, the information utilized by the detectors is limited within a fixed time interval, hence limiting the representative capacity of the learned features for classification. The similar problem is encountered in the aspect of localization. During training phase, \js{models could be supervised} 
	by only a temporal fragment of the tubes, which can output accurate local proposals but may fail to locate entire tubes in a consistent manner.
	\js{Additionally, IOU-based linking algorithms may result in accumulative localization error when noisy bounding box proposals are produced.} Since the 
	\js{transition within} action tubes is usually smooth and gradual, \js{we hypothesize that} using lesser number of boxes could be adequate to depict the action tube shape. 
	\js{Current pipelines, in their present state, relies heavily on dense per-frame predictions, which are redundant and a hindrance to efficient action detection.} 
	
	
	With these considerations, we depart from classic detect-and-link strategies by proposing a new coarse-to-fine action detector (CFAD) that can generate more accurate action tubes with higher efficiency. Unlike previous approaches that detect dense boxes at first, the CFAD (as illustrated in Fig.~\ref{fig:coarse-fine}) \js{goes on a progressive approach of estimating at a rougher level before ironing out the details}. This strategy first estimates coarser action tubes, and then selectively refine these tubes at key timestamps. The action tubes are generated via two important components in our pipeline: Coarse Module and Refine Module. 
	
	\lyx{The Coarse Module is designed to address the lack of global information and low efficiency in previous detect-and-link paradigm}. In a \emph{global} sense, it supervises the tube regression with the full tube shape information. \lyx{In addition}, within this module, a parameterized modeling scheme is introduced to depict action tubes. \lyx{Instead of predicting large amount of box location at each frame, Coasre Module only predict a few trajectory parameters to describe the tube of various endurance}. As a result, this module learns a \js{robust representation that accurately and efficiently characterizes action tube changes.}
	
	
	The Refine Module delves into the \emph{local} context of each tube, to find precise temporal locations that are essential to further improve the estimated action tubes, which in turn, improves overall detection performance and efficiency. To properly refine the action tubes, a \lyx{labelling} algorithm is designed to generate labels that guide the learning of key timestamps selection. By a search scheme, the original coarse boxes are replaced by the largest scoring box proposals at these temporal locations, which then interpolate the final tube.
	
	
	
	In summary, our contributions are three folds. (1) We propose a novel \emph{coarse-to-fine} framework for the task of spatial-temporal action detection, which \js{differs from} the \js{conventional} paradigm of detect-and-link. Our new pipeline achieves state-of-the-art results on standard benchmarks with inference speed of $3.3 \times$ faster than the nearest competitor. (2) Under this framework, we design a novel action tube estimation method based on parametric modeling to \lyx{fully exploit global supervision signal and} handle time variant box coordinates \lyx{by predicting limited amount of parameters}. (3) We also propose a simple yet effective method of predicting an importance score for each sampled frame which is used to select key timestamps for the \js{refinement of output} action \js{tubes}.
	
	
	\section{Related Works}
	\subsection{Action recognition}
	Deep learning techniques have \js{shown to be} effective and powerful in the classification of still images~\cite{he2016deep,huang2017densely,simonyan2014very}, and some existing works have extended such \js{schemes to} the task of human action recognition in video. \js{Direct extensions attempt to model sequential data with serial or parallel networks.}  ~\cite{li2018videolstm,sun2017lattice} combined 2D CNN with a RNN structure to model spatial and temporal relations separately. In ~\cite{simonyan2014two}, the authors found that the involvement of optical flow is beneficial for temporal modeling of actions and thus, proposed a two-stream framework that 
	extracts features from RGB and optical flow data \js{using separate parallel networks; the inference result being the combination of both modalities.} In \cite{Tran_2015}, the authors designed a 3D convolution architecture to automatically extract a high dimensional representation for input video. The I3D network \cite{carreira2017quo} further improved the 3D convolution technique by inflating convolution kernels of networks pre-trained on ImageNet (2D) \cite{deng2009imagenet} into an efficient 3D form for action recognition. Although these methods achieved good results on video classification benchmarks, they can only make video level predictions and are unable to ascertain 
	the position of actors and the duration of action instances.
	
	\subsection{Spatio-temporal action detection}
	The task of spatio-temporal action detection is more complex than direct classification of videos. It requires both correct categorization and accurate localization of actors during the time interval when the action happens. Gkioxiari \etal proposed the first pipeline for this task in~\cite{Gkioxari_2015}, where R-CNN~\cite{girshick2014rich} was applied on each frame to locate actors and classify actions, the results are then linked by viterbi algorithm. Saha \etal \cite{saha2016deep} designed a potential-based temporal trimming algorithm to extend general detection methods to untrimmed video datasets. Following the workflow of these two works,~\cite{peng2016multi,kalogeiton2017action,hou2017end,qiu2019learning} tried learning more discriminative features of action instances with larger spatial or temporal context, \js{a concept greatly enhanced by ~\cite{li2018recurrent} through a multi-channel architecture that learns recurrently from tubelet proposals.} 
	Some works~\cite{singh2017online,huang2018online} aimed to improve heuristic linking for better localization. Recent works ~\cite{zhao2019dance,su2019improving} proposed innovative two-stream fusion schemes for this task.~\cite{yang2019step} took a novel route to progressively regressing clip-wise action tubelets and linking them along time. Overall, all these works require temporally dense detections for each video, which is cumbersome. This inefficiency \lyx{gets worse} when optical flow computation is taken into account.
	
	Among the existing works,~\cite{yang2019step} is the most similar to CFAD with its refinement process. However, our method is different from it from three aspects. Firstly, CFAD estimates coarse level tubes with parametric modeling and global supervision, while~\cite{yang2019step} relies on per-frame detection. Secondly, our approach does not require further temporal linking or trimming process. Finally,~\cite{yang2019step} refines the boxes densely for each frame, while our method only refines the box locations at selected key timestamps.
	
	
	\subsection{Weight prediction}
	Weight prediction is a meta-learning concept where machine learning models are exploited to predict parameters of other structures~\cite{andrychowicz2016learning}. For example, the STN~\cite{jaderberg2015spatial} utilized deep features to predict affine transformation parameters. \cite{hu2018learning} used category-specific box parameters to predict an instance weighted mask, while MetaAnchor~\cite{yang2018metaanchor} learned to predict classification and regression functions from both box parameters and data. The Coarse Module of our method is also inspired by such similar mechanisms, \lyx{where the trajectory parameters are predicted by relevant \js{spatio-temporal} structures to depict the tube variation along time}. \js{To the best of our knowledge,} 
	our approach is the first \js{attempt} at exploiting parameterized modeling to handle action tube \js{estimation}.

	\begin{figure}[tb]
		\centering
		\includegraphics[width=0.9\textwidth]{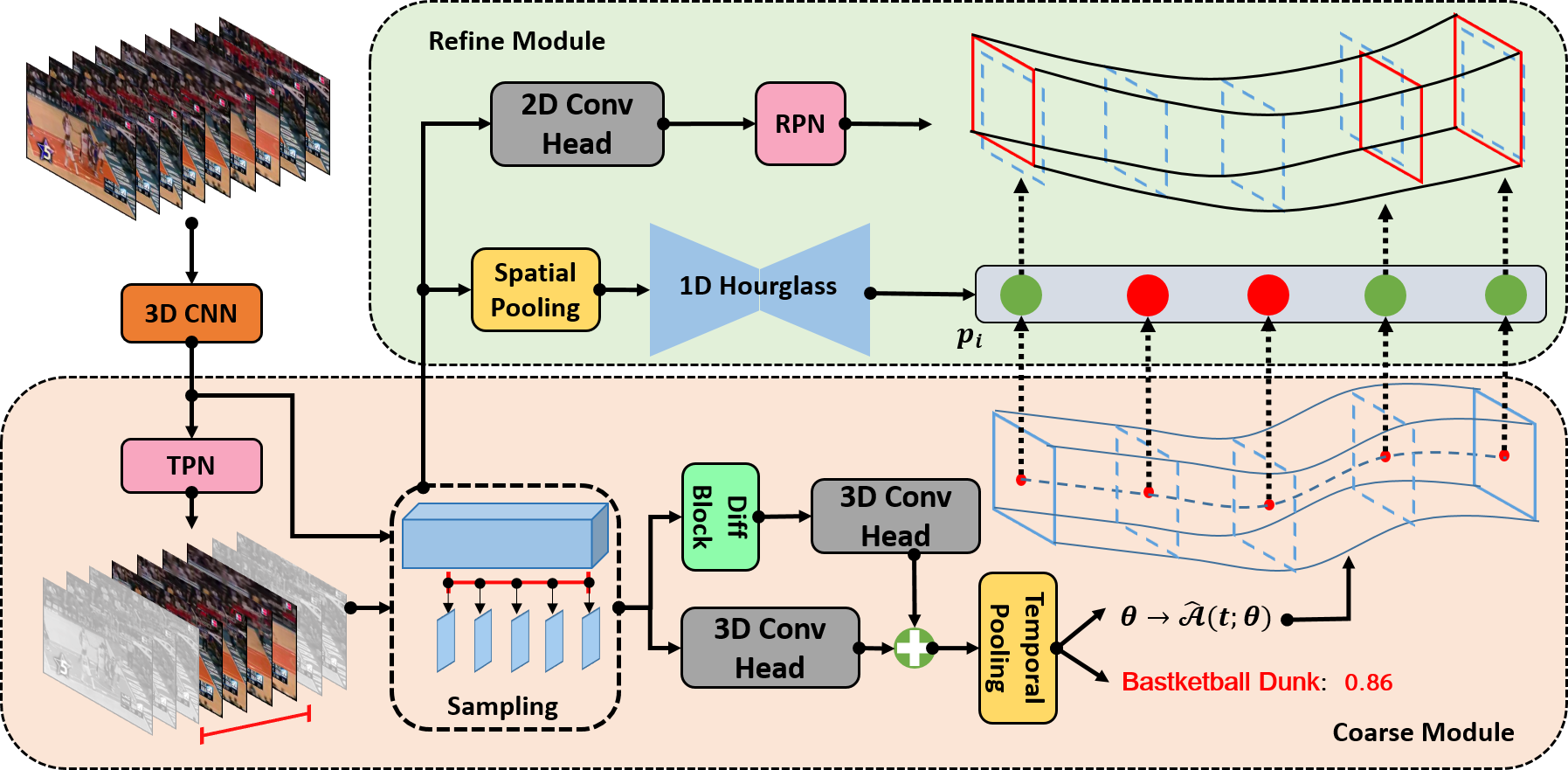}
		\caption{Overview of the proposed CFAD framework. TPN block denotes the temporal action proposal network. A 3D Conv Head block indicates a cascaded NL-3D Conv structure (``NL'' represents the NonLocal Block of~\cite{Wang_2018}). 2D Conv Head block denotes cascaded 2D spatial convolutions. (Best viewed in color)}\label{fig:overview}
		
	\end{figure}
	
	\section{Methodology}
	\subsection{Framework}
	In this section, we introduce the proposed Coarse-to-Fine Action Detector (CFAD) in detail. We first formulate the problem and provide an overview of our approach. Then we discuss more elaborately on the two primary components of CFAD -- the Coarse Module for tube estimation and Refine Module for final proposal.
	
	One action tube instance in videos can be formulated as a set, $\mca{A}=\{(t_i, b_i)|i=0,\cdots,T_A-1\}$, where $t_i$ is the timestamp of a certain frame, $b_i = (x_i, y_i, w_i, h_i)$ is the corresponding actor box within this frame, and $T_A$ denotes the total number of bounding boxes in a ground-truth tube. Each tube $\mca{A}$ is accompanied with a category label $c$. 
	
	The workflow of CFAD is shown in Fig.~\ref{fig:overview}. Firstly, the input video is \lyx{resampled to a fixted length $T$ and} fed into 3D CNN for spatio-temporal feature extraction. Then the feature is processed a temporal proposal network (TPN) to obtain class-agnostic temporal proposals \lyx{$(t_s, t_e)$. $t_s$ is the start timestamp and $t_e$ denotes the end timestamp}. In this paper, we instantiate the temporal proposal network by implementing one that is similar to~\cite{xu2017r}. Given the temporal proposal, we uniformly sample $N$ 2D features $\{\mbf{F}_i|i \in [0, N-1] \}$ along the time axis within interval $(t_s, t_e)$, which are sent to Coarse and Refine Module simultaneously. In Coarse Module, these 2D sampled features are used to estimate coarse level action tubes. Next, the estimated tube and the sampled 2D features in the Refine Module are exploited for frame selection and \js{tube refinement at identified key timestamps}. 
		
		\subsection{Coarse tube estimation}\label{sec:coarse}
		
		\lyx{We design two convolutional brunches in the Coarse Module to process} The sampled 2D features, one branch processes the input features directly and the other branch 
		handles the temporal residual component $\{\mbf{F}_{i+1}-\mbf{F}_i|i=0,N-2\}$ of the input features, which is output through the ``Diff Block'' in Fig.~\ref{fig:overview}. We add residual processing since the temporal residual component can provide more time variant information, which is beneficial to discriminate different actions and predict localization changes along time. For each branch, a Non-Local Block~\cite{Wang_2018} is cascaded with a 3D convolution blocks to construct the ``3D Conv Head'' module  in Fig.~\ref{fig:overview}, which aggregates information from both spatial and temporal context. The output of the two branches are fused by element-wise summation and 
		aggregated via temporal average pooling.
		
\begin{figure}[tb]
			\centering
			\subfigure[]{ \label{fig:estimate}\includegraphics[width=0.45\textwidth]{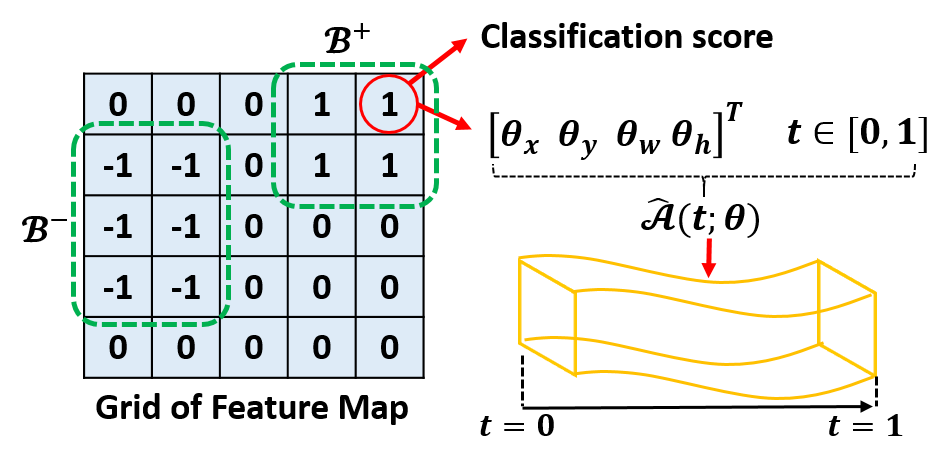}}
			\subfigure[]{ \label{fig:label}\includegraphics[width=0.45\textwidth]{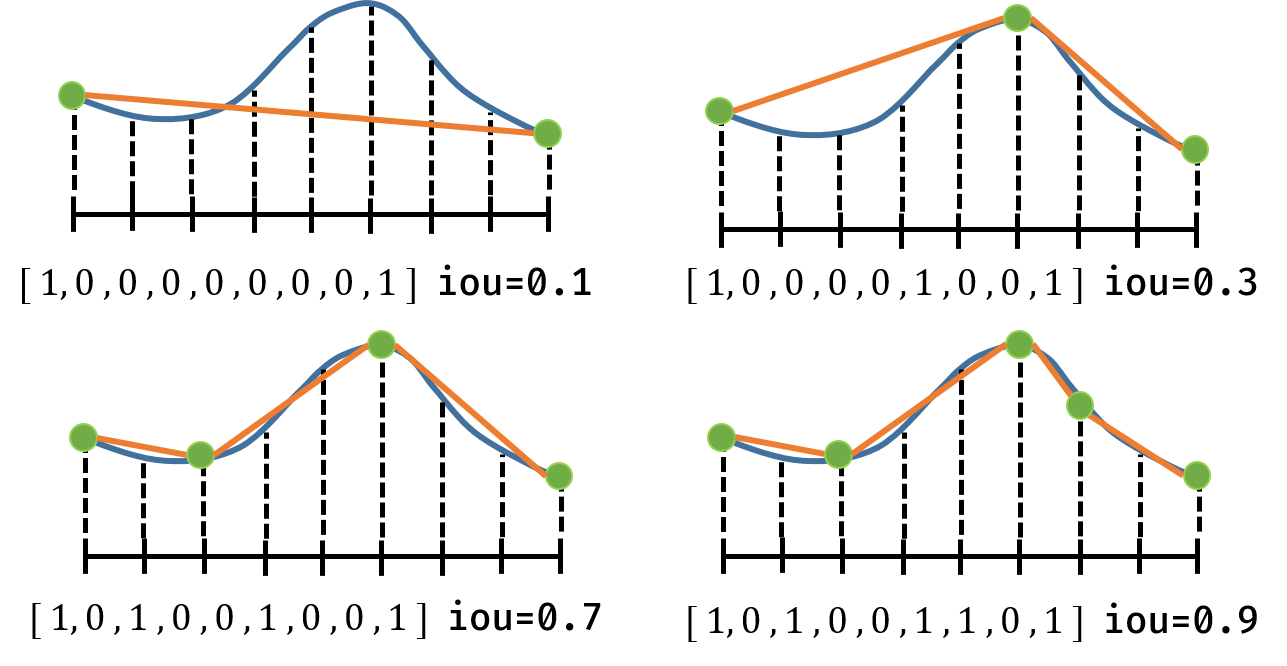}}
			
			\caption{(a). Illustration of coarse tube estimation, where ``1'' and ``-1'' symbols denote the positive and negative samples after matching, and ``0'' are ignored samples. (b) Key timestamps label selection process in the refine module. For ease of simplification, this figure depicts the case of 1-dimensional linear interpolation. The blue curve is the ground-truth, the orange one is the interpolated curve and green nodes represent selected timestamps. (Best viewed in color)}
			
		\end{figure}

		To estimate coarse-level action tubes, we adopt a \emph{parameterized modeling} scheme, where \lyx{we define a coarse-level tube estimation as a parameterized mapping $\hat{\mca{A}}(t; \boldsymbol{\theta}): [0,1]\longrightarrow \mathbb{R}^4$}. \lyx{$\hat{\mca{A}}$ tries to predict the coarse spatial location, \lyx{i.e. [x(t), y(t), w(t), h(t)]}, given a normalized timestamp $t$ and trajectory parameter $\boldsymbol{\theta}$.} The mapping parameters $\boldsymbol{\theta}$ are predicted by the deep features from the temporal pooling block. To this end, we slide predefined anchor boxes of different sizes on the 2D output feature map from temporal pooling block to obtain positive samples $\mca{B}^+$ and negative samples $\mca{B}^-$ as according to the IOUs between anchors and tubes (illustrated in Fig.~\ref{fig:estimate}). For each sample in $\mca{B}^+$, the network should predict its corresponding classification score and the tube shape parameter $\boldsymbol{\theta}$ \lyx{through an additional $1\times1$ convolution layer}.
		
		\textbf{Segment-wise matching.} \lyx{To measure the overlap between an anchor box $b_a$ and ground-truth $\mca{A}$, an intuitive idea is to calculate the average value of IOUs between $b_a$ and each boxes belonging to $\mca{A}$. However, since tube shapes \js{may} include motion and camera shake, such matching strategy might result in small IOU value and induce the imbalance issue of samples.} Hence, we design a segment-wise matching scheme to \lyx{separate positive and negative samples. To be specific, We take the boxes on first $K$ frames in $\mca{A}$ as a valid segment for matching positive anchors, where $K$ is a predefined segment length. We take the segment from the beginning of the tube because we found the final model performance is not sensitive to the segment position. If the average overlap between $b_a$ and these $K$ boxes is larger than a threshold, it is taken as a positive sample. Further, if $b_a$ has high overlap with multiple concurrent tubes, we choose the ground-truth with largest segment IOU as the matched tube. On the other hand, to find negative samples, we split the ground-truth tube $\mca{A}$ into $\floor{\frac{T_A}{K}}$ segments and compute IOUs between $b_a$ and each segments as discussed above, if the maximum IOU among all segments is still less than a threshold, then it is taken as a negative sample, the intuition behind such design is that negative samples should have low overlap with any boxes in $\mca{A}$}.
		
		
		\textbf{Parameterized modeling.}
		Generally, any parameterized 
		function that takes 
		\js{a single}
		scalar as input and outputs a 4-dimensional vector can be used as the tube mapping. In this paper, we use the family of high order polynomial functions to model action tube variations along the timestamp. This is because action tubes typically change smoothly and gradually, while polynomial functions are capable enough of describing the patterns of tube shape changes. Therefore, the instantiation of parameterized 
		estimation function $\hat{\mca{A}}(t;\boldsymbol{\theta})$ can be formulated as:
		\begin{equation}
		\hat{\mca{A}}(t;\boldsymbol{\theta}) = \left[x(t;\boldsymbol{\theta}_x), y(t;\boldsymbol{\theta}_y), w(t;\boldsymbol{\theta}_w), h(t; \boldsymbol{\theta}_h)\right] = \left[\boldsymbol{\theta}_x^T\boldsymbol{t},\boldsymbol{\theta}_y^T\boldsymbol{t},\boldsymbol{\theta}_w^T\boldsymbol{t},\boldsymbol{\theta}_h^T\boldsymbol{t}\right]
		\end{equation}
		where \lyx{the trajectory of each coordinate is regarded as a polynomial curve of order $k$}, the predicted parameter matrix $\boldsymbol{\theta}=[\boldsymbol{\theta}_x, \boldsymbol{\theta}_y, \boldsymbol{\theta}_w, \boldsymbol{\theta}_h]$ of size $(k+1)\times 4$ \lyx{is composed of the polynomial coefficient for each bounding box coordinates.} The vector $\boldsymbol{t}=[1, t, t^2, \cdots, t^k]^T$ \lyx{contains various orders of current timestamp}. To learn features that are invariant to anchor transitions, we do not use $\hat{\mca{A}}(t;\boldsymbol{\theta})$ to directly estimate the absolute coordinates, but instead perform estimation of relative coordinates w.r.t matched bounding box $b_a$ following the method of encoding in~\cite{ren2015faster}.
		
		During training, the model learns to separate positive and negative samples, to predict the correct action classes and relative coordinates of a coarse tube under the supervison of the loss function in Eq.~\ref{eq:coarse_loss},
		\begin{equation}\label{eq:coarse_loss}
		L_{coarse} = \frac{1}{|\mca{B}^+\cup\mca{B}^-|}L_c + \frac{1}{|\mca{B}^+|}L_r
		\end{equation}
		where $|\cdot|$ denotes the size of the set. $L_c$ is the classification loss in~\cite{ren2015faster} while $L_r$ is the regression loss from the supervision of the whole ground-truth tube:
		\begin{equation}\label{eq:enc}
		L_r = \frac{1}{T_A}\sum_{b_a \in \mca{B}^+}{\sum_{(t_i, b_i) \in \mca{A}}{\left|\left|\hat{\mca{A}}(\hat{t}_i;\boldsymbol{\theta}_a)-\textbf{enc}(b_i, b_a)\right|\right|^2}}
		\end{equation}
		The function $\textbf{enc}(\cdot, \cdot)$ in Eq. \ref{eq:enc} is the same as the encoding function in~\cite{ren2015faster} to \lyx{encode the 4-dimensional relative offsets from anchor box to ground-truth box}. $\boldsymbol{\theta}_a$ is the predicted tube shape parameter associated with anchor $b_a$. The symbol $\hat{t}_i$ defined in Eq. \ref{eq:normalized_ts} is the normalized timestamp of ground-truth bounding boxes in tube $\mca{A}$. We normalize the input timestamp before calculating the tube shape in order to avoid value explosion when the polynomial order \js{increases}. 
		\begin{equation}\label{eq:normalized_ts}
		\hat{t}_i = \frac{t_i-t_0}{t_{T_A-1}-t_0} \quad 
		\forall\ (t_i, b_i) \in \mca{A} 
		\end{equation}
		
		\subsection{Selective refinement}
		
		After the estimated coarse tube $\hat{\mca{A}}(t;\boldsymbol{\theta})$ has been generated by the Coarse Module, its location is further refined in the Refine Module.
		 The Refine Module first selects the samples attached with key timestamps for action tube localization, and then refine tube boxes based on these features \lyx{and guidance of coarse tube}. 
		
		\textbf{Key timestamp selection.} One simple and intuitive refinement scheme is to observe the tube location at each sampled 2D feature map and then refine the box according to the the features within that area. However, when the sample number $N$ increases, \js{such a scheme is costly in computation.} Since changes in the action tubes are usually smooth, there is only a limited number of \js{sparsely distributed} bounding boxes that are decisive to the shape of tubes. 
		Thus, we design a selector network in the Refine Module to dynamically sample key timestamps that are most essential for 
		localization.
		
		In our implementation, we perform importance evaluation by squeezing the input \lyx{2D sampled features $\{\mbf{F}_i|i \in [0, N-1]\}$} with spatial pooling and applying a 1D hourglass network along the time dimension. This outputs an importance score $p_i$ for each sample (shown in Fig.~\ref{fig:overview}). During inference phase, we only take samples that satisfy $p_i \geq \alpha$ as samples of key timestamps and then proceed to refinement.
		
		\lyx{In the training phase}, we heuristically define sets of labels to guide the training of the selector network. Specifically, first the ground-truth action tube $\mca{A}$ is uniformly split into $N-1$ segments along temporal axis with $N$ endpoints. The $i$-th endpoint is associated with the $i$-th sampled feature $\mbf{F}_i$, and its normalized timestamp is defined as $s_i=i/(N-1)$. Let the timestamp set be defined as $\mca{U}=\{s_i|i=0,\cdots,N-1\}$ and the key points set as $\mca{U}_k$. 
		We start from $\mca{U}_k=\{s_0, s_{N-1}\}$ having the start and end points and gradually append other $s_i$ into $\mca{U}_k$. The process can be illustrated in Fig.~\ref{fig:label}, whereby for each iteration, we greedily select the timestamps $s^*$ which maximizes the overlap between the interpolated tube and ground-truth $\mca{A}$ as in Eq.~\ref{eq:label}, and append this timestamp into $\mca{U}_k$. The process stops when the IOU between interpolated tube and ground-truth tube is larger than a predefined threshold $\epsilon$.
		\begin{equation}\label{eq:label}
		s^* = \arg\max_{s_i \in \mca{U}/\mca{U}_k}{IOU\left(\textbf{Interp}(\mca{U}_k\cup\{s_i\}), \mca{A}\right)}
		\end{equation}
		Here, the function $\textbf{Interp}(\cdot)$ can be any polynomial interpolation over the input timestamp set. To avoid the \lyx{large numerical oscillation around the endpoint}, we choose the piece-wise cubic spline interpolation in this paper as instantiation. We assign feature samples 
		in $\mca{U}_k$ with label $1$ and samples in $\mca{U}/\mca{U}_k$ with label $0$. \js{We utilize} these labels 
		to train the timestamp selector network with binary cross-entropy loss.
		
		\textbf{Sample-wise location refinement.} In the Refine Module, the selected 2D features are first processed by cascaded 2D convolution blocks (shown in Fig.~\ref{fig:overview}), then a class-specific regional proposal network (RPN)~\cite{ren2015faster} is applied over these features to generate bounding box proposals at corresponding timestamps. With the estimated action tube function $\hat{\mca{A}}(t;\boldsymbol{\theta})$, we can now obtain the estimated action bounding boxes at $i$-th sampled timestamps $s_i$ with Eq.~\ref{eq:sample}, where $\textbf{dec}(\cdot)$ 
		is the \lyx{inverse operation of $\textbf{enc}(\cdot, \cdot)$ in Eq.~\ref{eq:enc}}.
		\begin{equation}\label{eq:sample}
		\hat{x_i},\hat{y_i},\hat{w_i},\hat{h_i} = \textbf{dec}\left( \hat{\mca{A}}(s_i;\boldsymbol{\theta})\right)
		\end{equation}
		\noindent We design a simple local search scheme to refine the estimated bounding box at selected key timestamps. For each selected 2D sample, a searching area $\Omega$ is defined as,
		\begin{equation}\label{eq:area}
		\Omega = [\hat{x_i}-\sigma\hat{w_i},\hat{x_i}+\sigma\hat{w_i}] \times [\hat{y_i}-\sigma\hat{h_i},\hat{y_i}+\sigma\hat{h_i}]
		\end{equation}
		where $\sigma$ is a hyperparameter that controls the size of searching area. We choose the action box proposal (from RPN) with the largest score where its center is located inside $\Omega$, as the replacement of the original \js{coarsely} estimated box.
		
		The final output action tube is obtained via interpolation over all refined \lyx{boxes} and unrefined bounding boxes (\lyx{localized via Eq.~\ref{eq:sample}}). The associated action score is the smooth average of classification score and RPN score.
		
		\section{Experiment Results}
		\subsection{Experiment configuration}
		
		\noindent \textbf{Datasets.} We conduct our experiment on \lyx{three} common datasets for the task of action tube detection -- UCF101-24, \lyx{UCFSports} and JHMDB-21 datasets. \lyx{Although the AVA \cite{Gu2018AVAAV} dataset also includes bounding box annotations, it mainly focuses on the problem of \emph{atomic action classification} on sparse single key frames instead of \emph{spatiotemporal action detection} at the video level, which is the task we are focusing here. Hence, we did not conduct our experiments on the AVA dataset.}
		
		The \emph{UCF101-24 dataset}~\cite{soomro2012ucf101} contains 3,207 untrimmed videos with frame level bounding box annotations for 24 sports categories. The dataset is challenging due to frequent camera shake, dynamic actor movements and a large variance in action duration. Following previous works~\cite{saha2016deep}, we report results for the first split with 2,275 videos for training and the other videos for validation. \lyx{We use the corrected annotation~\cite{singh2017online} for model training and evaluation}. The \emph{JHMDB-21} is a subset of HMDB-51 dataset~\cite{Jhuang2013}, which contains a total of 928 videos with 21 types of actions. All video sequences are temporally trimmed. The results are reported as the average performance over 3 train-val splits. \lyx{The \emph{UCFSports} dataset~\cite{ucfsports} contains 150 trimmed videos in total and we report the results on the first split.} Note that although videos in JHMDB-21 and UCFSports are trimmed temporally, their samples are still suitable for our framework as they comprised mostly of cases where actions span the whole video.
		
		\smallgap\textbf{Metrics.} We report the video-mAP (v-mAP)~\cite{Gkioxari_2015} with different IOU thresholds as our main evaluation metric for spatial-temporal action localization on all datasets. \lyx{In addition, frame-level mAP at threshold $0.5$ is reported to evaluate per-frame detection performance.} A proposal is regarded as positive when its overlap with 
		the ground-truth is larger than threshold $\delta$. We also adopt video-level mean Average Best Overlap (v-MABO)~\cite{kalogeiton2017action} in the ablation study to evaluate the localization performance of our approach. The criterion calculates the mean of largest overlap between ground-truth tubes and action proposals, \js{averaged over all classes.}
		
		\smallgap\textbf{Implementation details.}
		We use the I3D network \cite{carreira2017quo} pretrained on Kinetics-600 as our 3D feature extractor, taking the feature from \emph{mixed\_5b} layer as our 3D feature. We set the video resampling length $T$ to $96$ frames for UCF101-24 and $32$ frames for JHMDB-21 and UCFSports. The hyperparameter $\epsilon$ in our paper is set to $0.8$ and the segment length for matching $K$ is set to $6$ frames. The number of sampling points $N$ is set to $16$ for UCF101-24, $6$ for JHMDB-21 and $8$ for UCFSports according to the average length of action instances. For the anchor design, we follow the strategy of~\cite{redmon2017yolo9000} by clustering the bounding boxes from training set into 6 centers and taking their respective center coordinates as the default anchor boxes. In the training phase, the temporal proposal network is trained first, and then the \js{entire} network is trained end-to-end with temporal proposals and ground-truth span. 
		to learn the final action tubes.
		We use the SGD solver to train CFAD with a 
		batch size of $8$. In inference stage, to handle concurrent action instances, the Coarse Module outputs at most $3$ (the maximum number of instances in a video based on the datasets) estimated tubes followed by a tube-wise non-maximal suppression process with IOU threshold of $0.2$ in order to avoid 
		duplicated action tubes.
		

		\subsection{Ablation study}
		In this section, we report the \lyx{video-mAP} results with $\delta=0.5$ of various ablation study experiments. The input modality is only RGB data unless specified.
		
		\begin{table}[tb]
			\centering
			\setlength{\tabcolsep}{3mm}{
				\begin{tabular}{c|cccc|ccc}\hline
					& \multicolumn{4}{c|}{UCF101-24} & \multicolumn{3}{c}{JHMDB-21} \\ \hline
					k & 2 & 3 & 4 & 5 & 1 & 2 & 3 \\ \hline
					no refine & 46.0 & 48.4 & 51.6 & 50.1 & 79.7 & 80.9 & 80.3 \\ \hline
					$\sigma=0.4$ & 57.5 & 57.6 & 58.8 & 58.0 & 80.8 & 82.5 & 81.3 \\ 
					$\sigma=0.6$ & 59.9& 60.1 & 61.7 & 60.0 & 81.4 & 83.2 & 82.4 \\
					$\sigma=0.8$ & 60.3 & 62.0 & \textbf{62.7} & 61.6 & 82.3 & \textbf{83.7} & 83.2\\\hline
				\end{tabular}
				\caption{Ablation study on the effectiveness of refinement with different hyperparameter 
				\js{settings.}
				}
				\label{tab:parameters}}
		\end{table}
		
		\begin{figure}[tb]
			\centering
			
			\subfigure[]{\label{fig:mABO}\includegraphics[width=0.49\textwidth, height=0.18\textheight]{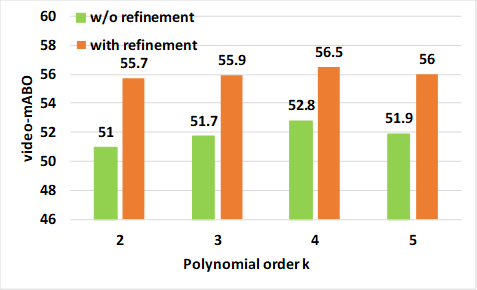}}
			\subfigure[]{\label{fig:select}\includegraphics[width=0.49\textwidth, height=0.18\textheight]{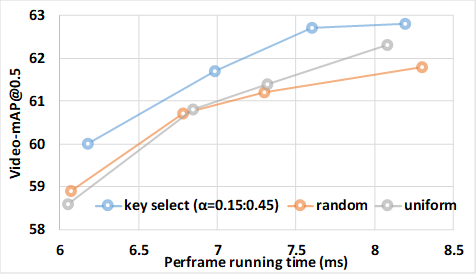}}
			\caption{(a). v-MABO value of action tubes with different polynomial orders on UCF101-24. (b) Time-performance trade-off with different timestamp selection schemes on UCF101-24. (best viewed in color)}
			\label{fig:plot}
			
		\end{figure}
		
		\textbf{The effectiveness of refinement}. First, we analyze the \js{effects of the refinement process} on the accuracy of coarse tube estimation \lyx{on UCF101-24 and JHMDB-21}. The results are reported in Table~\ref{tab:parameters} where ``no refine'' denotes the configuration without location refinement. From these results, it is obvious that the Refine Module can bring large improvements in v-mAP regardless of the polynomial order of estimated tubes; the largest performance gain can be up to $+14.3\%$ when $k=2$ on UCF101-24. The improvement is less obvious on JHMDB-21, which we think is owing to the fact that JHMDB-21 is less dynamic and coarse-level estimations \js{may be} 
		close to the ground-truth tubes. We also evaluate the v-MABO value on UCF101-24 as shown in Fig.~\ref{fig:mABO}, where improvements  
		by at most $+4.7\%$ \js{are possible} by the refinement process. The results show that the Refine component is essential to better detection performance. 
		
		Meanwhile, from Table~\ref{tab:parameters}, we can also see that as the searching area gets larger, the mAP performance can be improved to some extent, since larger searching area can cover more centers of action proposals. We did not try larger searching area \emph{i.e.}~$\sigma > 0.8$ since we find the performance improvement is marginal (less than $+0.2\%$) \js{beyond $\sigma=0.8$.} This is because larger searching area also makes the refinement more vulnerable to noisy proposals. 
		
		\textbf{Polynomial order selection}. We also report in Table~\ref{tab:parameters} the effect of different polynomial order $k$ which decides the form of estimated tube mapping $\hat{\mca{A}}(t;\boldsymbol{\theta})$. Overall, we find that the performance improves along with the increase of $k$ \js{for both with and without refinement}, since higher order polynomial functions show stronger ability in characterizing variations of action tubes. 
		
		On the other hand, we found that \js{as the order gets} larger (than $k=5$ on UCF101-24 and $k=3$ on JHMDB-21), the detection performance drops comparatively against the optimal value in both cases. We think the reason behind this is that although higher order polynomial functions are usually more representative, they are more complex requiring more coefficients, and the parameters predicting coefficients of higher orders are more difficult to be trained efficiently since the corresponding gradients are very small. The similar tendency is also reflected in the MABO results on UCF101-24 shown in Fig.~\ref{fig:mABO}, where the localization did improve (for both refine and no refine cases) from $k=2$ to $k=4$, but MABO drops after that with higher orders. Also, we observed during training that configurations with a larger polynomial order tends to slow down the convergence process and possibly result in numerical oscillation of the loss function. 
		
		\begin{figure}[tb]
			\centering
			\includegraphics[width=0.98\textwidth]{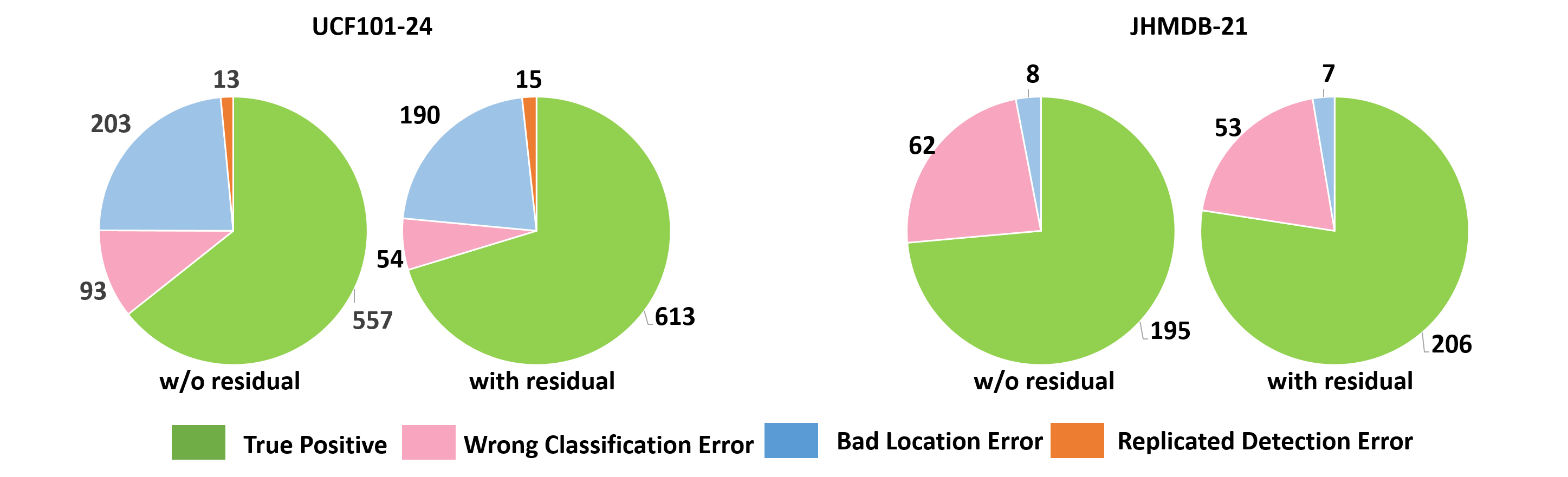}
			\caption{Statistics of \js{true positive and} various false positive proposals of CFAD on UCF101-24 and split-3 of JHMDB-21. (Best viewed in color)}
			\label{fig:false}
			
		\end{figure}
		
		\textbf{Effectiveness of residual processing branch}. Here, we conduct experiments on UCF101-24 \lyx{and JHMDB-21} to analyze how the temporal residual information 
		impacts the output action tube results. \js{To test this}, we remove the branch with differential module (``Diff Block'' in Fig.~\ref{fig:overview}) as our baseline. For a detailed comparison, we 
		\js{break down the final} proposals into four mutually exclusive types.
		\lyx{
			\begin{itemize}
				\item \textbf{True Positive}: the proposal classifies an action correctly and has tube-overlap with ground-truth that is larger than $\delta$.
				\item \textbf{Wrong Classification Error}: a proposal with incorrectly classified action although it overlaps more than $\delta$ with ground-truth.
				\item \textbf{Bad Localization Error}: a proposal that has correct action class but it overlaps less than $\delta$ with ground-truth.
				\item \textbf{Duplicated Detection Error}: a proposal with correct action class and overlaps more than $\delta$ with a ground-truth that has been detected.
			\end{itemize}
			Fig.~\ref{fig:false} illustrates the statistics of these proposals with/without the differential module.} From Fig.~\ref{fig:false}, we observe that the residual processing branch is particularly important for \js{accurate} action classification. With the help of information from the temporal residual feature, wrongly-classified samples are reduced by $42\%$ \lyx{on UCF101-24 and $14\%$ on JHMDB-21}. Furthermore, models with residual processing also \js{benefit from better tube localization} while the overall recall also improves due to the increase in true positive results. These results are evidential of the effectiveness of temporal residual component in Coarse Module.
		
		\textbf{Key timestamp selection}. We conduct an experiment on UCF101-24 to analyze the impact of the proposed key timestamp selection mechanism. In the experiment setting, we gradually increase the selection threshold $\alpha$ from $0.15$ to $0.45$ (in increments of $0.1$) and report their respective v-mAP value and per-frame time cost. For comparisons, we design two baseline methods: (1) Random selection of samples from the input $N$ 2D features with their corresponding timestamps taken as key timestamps, denoted as ``random''. (2) Selection of timestamps across $s_i$ based on a fixed time step, denoted as ``uniform''.
		
		The time-performance trade-off curves are shown in Fig.~\ref{fig:select}. We can observe that when the per-frame time costs are similar, our dynamic selection scheme is superior to the other two baseline methods. It is also worth noting that when the time cost gets smaller, the performance of ``random'' and ``uniform'' deteriorates faster than our scheme. This result indicates that the key timestamp selection process finds the important frames for location refinement and is reasonably robust to the reduction of available 2D features.
		
		\subsection{Comparison with state-of-the-art}
		
		\begin{table}[t!]
			\small
			\centering
			\setlength{\tabcolsep}{1.0mm}{
				\begin{tabular}{c|cccc|cccc|cc}\hline
					method & \multicolumn{4}{c|}{JHMDB-21} & \multicolumn{4}{|c}{UCF101-24} & \multicolumn{2}{|c}{UCFSports} \\\hline
					$\delta$& 0.2 & 0.5 & 0.75 & 0.5:0.95 & 0.2 & 0.3 & 0.5 & 0.5:0.95 & 0.2 & 0.5 \\\hline\hline
					\multicolumn{11}{l}{\textbf{2D backbone}} \\ \hline
					Saha \etal \cite{saha2016deep}& 72.6  & 71.5& - & - & 66.7 & 54.9 & 35.9 & 14.4 & - & - \\\hline
					Peng \etal \cite{peng2016multi}& 74.3 & 73.1 & - & - & 72.8 & 65.7 & 30.9 & 7.1 & 94.8 & 94.7 \\\hline
					Saha \etal \cite{saha2017amtnet} & 57.8 & 55.3 & - & - & 63.1 & 51.7 & 33.0 & 10.7 & - & - \\\hline
					Kalogeiton \etal \cite{kalogeiton2017action} & 74.2 & 73.7 & 52.1 & 44.8 & 76.5 & - & 49.2 & 23.4 & 92.7 & 92.7 \\\hline
					Singh \etal \cite{singh2017online} & 73.8 & 72.0 & 44.5 & 41.6 & 73.5 & - & 46.3 & 20.4 & - & -  \\ \hline
					Yang \etal \cite{yang2019step} & - & - & - & - & 76.6 & - & - & - & - & - \\\hline
					Zhao \etal \cite{zhao2019dance} & - & 58.0 & 42.8 & 34.6 & 75.5 & - & 48.3 & 23.9 & - & 92.7 \\\hline
					Rizard \etal \cite{Pramono_2019_ICCV} & 86.0 & 84.0 & 52.8 & 49.5 & 82.3 & - & 51.5 & 24.1 & - & - \\\hline
					Song \etal \cite{song2019tacnet} & 74.1 & 73.4 & 52.5 & 44.8 & 77.5 & - & 52.9 & 24.1 & - & - \\\hline
					Li \etal \cite{li2018recurrent} & 82.7 & 81.3 & - & -  & 76.3 & 71.4 & - & - & \textbf{97.8} & \textbf{97.8} \\ \hline
					Li \etal \cite{li2020actions} & 77.3 & 77.2 & \textbf{71.7} & \textbf{59.1} & 82.8 & - & 53.8 & \textbf{28.3} & - & - \\ \hline\hline
					\multicolumn{11}{l}{\textbf{3D backbone}}  \\ \hline
					Hou \etal \cite{hou2017end}  & 78.4 & 76.9 & - & - & 73.1 & 69.4 & - & - & 95.2 & 95.2\\ \hline
					Gu \etal \cite{Gu2018AVAAV}  & - & 76.3 & - & - & - & - & 59.9 & - & - & -\\ \hline
					Su \etal \cite{su2019improving} & 82.6 & 82.2 & 63.1 & 52.8 & \textbf{84.3} & - & 61.0 & 27.8 & - & - \\ \hline
					Qiu \etal \cite{qiu2019learning}  & 85.7 & 84.9 & - & - & 82.2 & 75.6 & - & - & - & - \\ \hline
					CFAD  & 84.8 & 83.7 & 62.4 & 51.8 & 79.4 & 76.7 & 62.7 & 25.5 & 90.2 & 88.6\\\hline
					CFAD\textbf{+OF} & \textbf{86.8} & \textbf{85.3} & 63.8 & 53.0 & 81.6 & \textbf{78.1} & \textbf{64.6} & 26.7 & 94.5 & 92.7 \\\hline
					
				\end{tabular}
			}
			\caption{Comparison with state-of-the-art methods (on video-mAP), `-' denotes that the result is not available, `\textbf{+OF}' indicates the input is combined with optical flow. All compared methods take both RGB and optical flow as input except~\cite{saha2017amtnet,hou2017end}}
			\label{tab:sota}
			
		\end{table}
		
		\begin{table}[tb]
			\centering
			\setlength{\tabcolsep}{2.0mm}
			\begin{tabular}{c|c|c}\hline
				method & input modal & frame-mAP@0.5 \\\hline
				Peng \etal \cite{peng2016multi} & RGB+OF & 65.7 \\
				Kalogeiton \etal \cite{kalogeiton2017action}& RGB+OF & 69.5 \\
				Yang \etal \cite{yang2019step} & RGB+OF &75.0 \\
				Rizard \etal \cite{Pramono_2019_ICCV}& RGB+OF &73.7 \\
				Song \etal \cite{song2019tacnet}& RGB+OF & 72.1 \\
				Gu \etal \cite{Gu2018AVAAV} & RGB+OF & \textbf{76.3} \\
				CFAD & RGB+OF & 72.5 \\\hline
				Hou \etal \cite{hou2017end} & RGB & 41.4 \\
				Yang \etal \cite{yang2019step} & RGB & 66.7 \\
				CFAD & RGB & \textbf{69.7} \\\hline
			\end{tabular}
			\caption{Comparison with state-of-the-art methods on frame-level mAP@0.5 on UCF101-24 dataset. `\textbf{+OF}' indicates the input is combined with optical flow.}
			\label{tab:frame}
		\end{table}

		In this section, we compare the proposed CFAD with other recent state-of-the-art approaches in the spatio-temporal action localization task on the UCF101-24, JHMDB-21 and UCFSports benchmarks. These results are listed in Table~\ref{tab:sota}. We also evaluate the performance of CFAD with two-stream input, where the optical flow is extracted using the method of~\cite{brox2004high}. For simplicity, we opt for an early fusion strategy~\cite{zhao2019dance} to maintain efficiency of our approach.
		
		It is worth noting that in Table \ref{tab:sota}, our method with only RGB input outperforms most other approaches that rely on two-stream features \lyx{on UCF101-24 and JHMDB-21}. \lyx{While it is still worse than the state-of-the-art method on UCFSports, we think the reasons behind this can be that this dataset is relatively simpler and smaller in scale with less dynamic movements, thus it could be more challenging to learn robust tube estimation.} \js{For fair benchmarking, we compare our method with other approaches utilizing 3D spatiotemporal features}~\cite{hou2017end,Gu2018AVAAV,qiu2019learning,su2019improving}.
		With RGB as input, CFAD achieves competitive performance on all datasets under different tested threshold criterion. Overally, our method achieves state-of-the-art under small threshold while there is still a margin towards the performance of ~\cite{su2019improving,li2020actions} under more strict criterion. Besides, we also observe that the optical flow information is helpful for the overall detection performance.
		
		\begin{figure}[tb]
			\centering
			\subfigure[]{\label{fig:time}\includegraphics[width=0.45\textwidth]{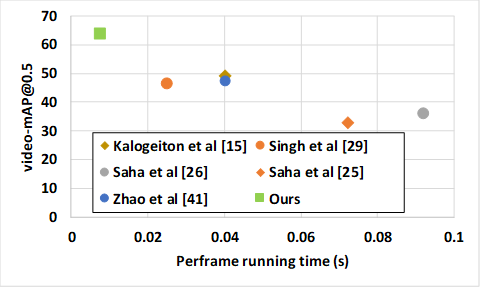}}
			\subfigure[]{\label{fig:param}\includegraphics[width=0.45\textwidth]{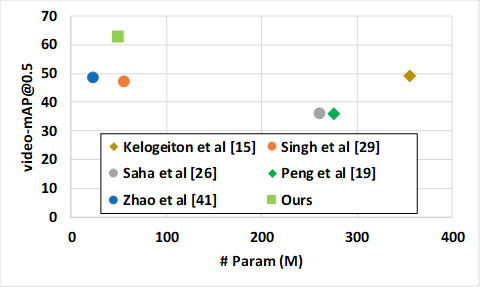}}
			\caption{(a). Comparisons of time-performance trade-off among different state-of-the-art approaches. (b). Comparisons of trade-off between model size and performance among different state-of-the-art approaches. (Best viewed in color)}
			
		\end{figure}
		
	    \textbf{Frame-mAP.} In Table~\ref{tab:frame}, we compare CFAD with other approaches on frame-level detection in UCF101-24. In our setting, we assign the video level score of a tube proposal to all boxes included by the tube to generate frame-level proposals. It can be observed that CFAD outperforms three pipelines with two-stream input. While it is still worse than some approaches \cite{Gu2018AVAAV,Pramono_2019_ICCV,yang2019step}, we think this is due to the less accurate interpolated boxes between sampled frames, which might result in many false positives with high score (which in turn lowers the overall metric). We argue that although such interpolation sacrifices frame-level accuracy, it enhances the system efficiency and video-level accuracy in return.  
		
		\textbf{Efficiency.} We also compare the runtime (inference) \lyx{and model size} of CFAD with RGB input on UCF101-24 against other approaches that also report their runtime. The speed is evaluated based on per-frame processing time, which is obtained by taking the runtime per video and dividing it by input length $T$. Since some other works only reported per-video time on JHMDB-21~\cite{saha2016deep,saha2017amtnet}, we compute their per-frame time in the same manner. The runtime comparison is illustrated in Fig.~\ref{fig:time} and \lyx{the model size comparison is reported in Fig.~\ref{fig:param}}. We observe that CFAD \lyx{only requires a small number of parameters (close to~\cite{singh2017online,zhao2019dance}, and much less than others) while} achieving superior running speed compared to other state-of-the-art methods. This vast improvement in processing efficiency can be attributed to the coarse-to-fine paradigm of CFAD, which does not require dense per-frame action detection followed by linking, and \lyx{the RGB input of CFAD avoids the additional computation to process optical flow.} Specifically, the proposed CFAD runs $\approx 3.3 \times$ faster than the nearest approach~\cite{singh2017online} ($7.6$ \emph{ms} vs. $25$ \emph{ms}). 
		
		\subsection{Qualitative results}
		
		\begin{figure}[tb]
			\centering
			\includegraphics[width=0.96\textwidth]{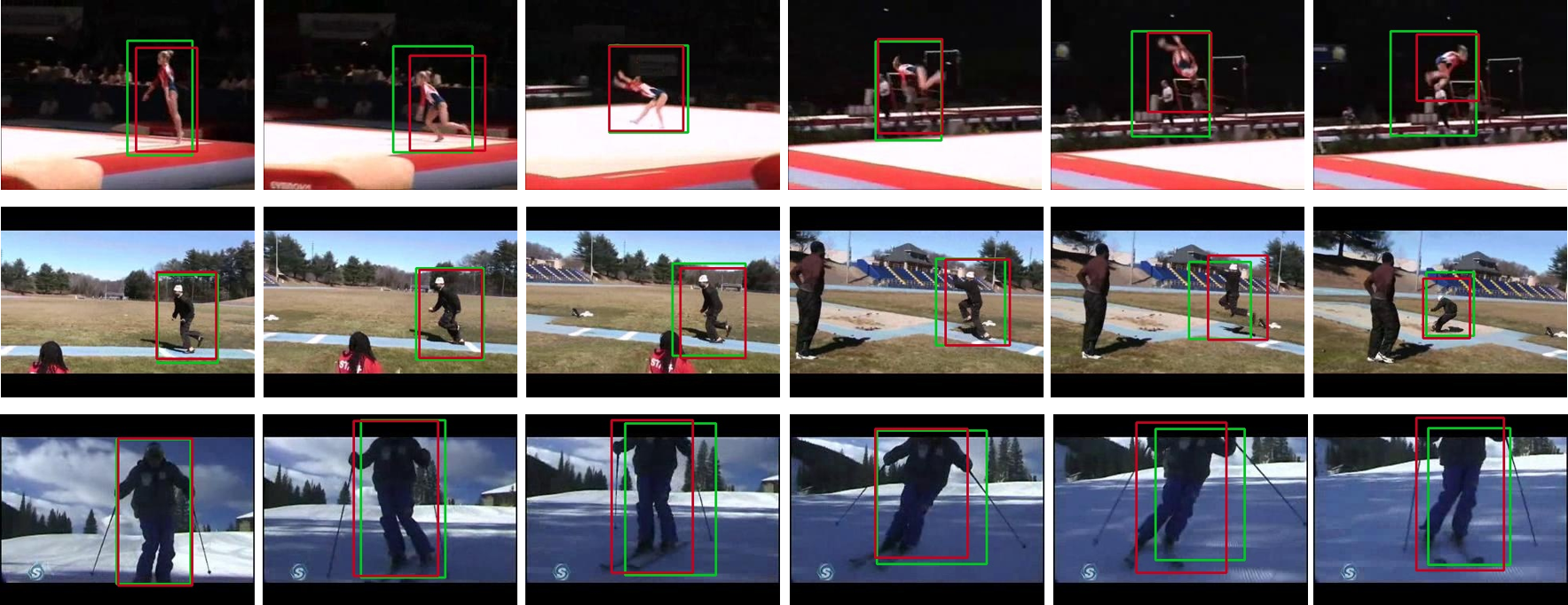}
			\caption{Visualization of detected action tubes. The green boxes denote the estimated action tubes from the Coarse Module. The red boxes are the final refined action tubes. (Best viewed in color)}
			\label{fig:vis}
		\end{figure}
		
		
		Fig.~\ref{fig:vis} shows some qualitative results of detected action tubes from the UCF101-24 dataset. The green boxes denote the estimated action tube output from the Coarse Module while the red boxes are the refined action tubes. We can observe that the selective refinement process has effectively corrected some poorly located action tubes, causing the bounding boxes \js{to wrap} tighter and more accurately around the actors. 
		These visuals can evidently explain the robustness of the coarse tube estimation method, and its capability at handling a variety of dynamic actions.
		
		\section{Conclusion}
		In this paper, we propose a novel framework CFAD for spatio-temporal action localization. Its pipeline follows a new coarse-to-fine paradigm, which does away with the need for dense per-frame detections. The action detector comprises of two 
		components (Coarse and Refine Modules) which play vital roles in coarsely estimating and then refining action tubes based on selected timestamps. Our CFAD achieves state-of-the-art results for a good range of thresholds on benchmark datasets and is also an efficient pipeline, running at  
		$3.3 \times$ faster than the nearest competitor.
		
 		\section*{Acknowledgement}
 		 The paper is supported in part by the following grants: China Major Project for New Generation of AI Grant (No.2018AAA0100400), National Natural Science Foundation of China (No. 61971277). The work is also supported  by funding from Clobotics under the Joint Research Program of Smart Retail.
		\bibliographystyle{splncs04}
		\bibliography{ref}
	\end{document}